# Effect of Degree Distribution on Evolutionary Search

Susan Khor

## ABSTRACT

This paper introduces a method to generate hierarchically modular networks with prescribed node degree list and proposes a metric to measure network modularity based on the notion of edge distance. The generated networks are used as test problems to explore the effect of modularity and degree distribution on evolutionary algorithm performance. Results from the experiments (i) confirm a previous finding that modularity increases the performance advantage of genetic algorithms over hill climbers, and (ii) support a new conjecture that test problems with modularized constraint networks having heavy-tailed right-skewed degree distributions are more easily solved than test problems with modularized constraint networks having bell-shaped normal degree distributions.

## Categories and Subject Descriptors

I.2.8 [**Artificial Intelligence**]: Problem Solving, Control Methods and Search.

## 1. INTRODUCTION

Many real-world networks of natural and man-made phenomena exhibit topological properties atypical of classical random graphs [1, 13]. Intra-organism networks e.g. networks of gene-regulation, metabolism and protein-protein interaction, exhibit broad connectivity of the kind that, allowing for finite sizes of the networks, exhibits a power-law decay, i.e. $P(k) \sim k^{-\gamma}$, $\gamma > 1$ where $P(k)$ is the probability that a randomly selected node is linked to $k$ other nodes and $\gamma$ is the degree exponent or scaling factor. For real-world networks, typically $2 \leq \gamma \leq 3$. Biological networks also exhibit higher than expected levels of *network clustering* or *transitivity* than comparable random networks [15]. Transitivity refers to the cliquishness of a network. Hierarchical organization is proposed as the key to combine a heavy-tailed right-skewed degree distribution with high network clustering within a single network, e.g. the *hierarchical network* model [16].

The objective in this paper is to explore the significance of hierarchical modularity and broad degree connectivity for evolutionary algorithms. Specifically, are problems whose constraint networks have heavy-tailed right-skewed degree distributions easier for evolutionary algorithms to solve? The *constraint network* of a problem specifies dependencies between parts of a problem.

To proceed with the investigation, a method to generate random hierarchically modular networks with a given list of node degrees drawn from a distribution is presented in section 2. A *modular network* has identifiable subsets of nodes with a higher density of links amongst nodes within a subset than between nodes of different subsets [14, 16]. A *hierarchically modular network* is one where the nodes can be recursively subdivided into *modules* (subsets of unexpectedly densely linked nodes) over several scales until some atomic level is reached [17]. Modules need not be isolated from one another, but are interrelated subsystems of a larger encompassing whole [17].

The significance of hierarchical modularity to the effectiveness of crossover in genetic algorithms [9] has been previously investigated [19]. That study demonstrated how in conjunction with tight genetic linkage, hierarchical modularity organizes the connectivity of problem variables in accordance with the *building-block hypothesis* [8] and thus produce test problems, e.g. the H-IFF problem class, which are more easily solved by genetic algorithms than hill climbers. The connectivity of the H-IFF problems is uniformly distributed and heterogeneously weighted. The constraint or inter-dependency network of H-IFF is fully connected and the links are unequally weighted. Hierarchical modularity in the H-IFF networks is the result of careful link weight arrangement – heavier weights are placed on intra-module than inter-module, and lower-level than higher-level links. Subsequent work also followed this connectivity and weight pattern [6]. The test problems generated in this paper depart from this mould – their constraint networks have heterogeneous connectivity and the links are equally weighted.

Like the *hierarchical random graph* model [4], our method to generate hierarchically modular networks uses a pre-specified topology to outline the hierarchical structure and guide the formation of modules. However, unlike the hierarchical random graph model and network generating models with inhomogeneous link probabilities [3], our method does not require a set of pre-determined link probability values. Instead, it utilizes the topology to determine *relatedness* between pairs of nodes in terms of *edge distances*. Nodes that belong to the same module are more related to each other than nodes that belong to different modules. The modularization algorithm performs link switching that favours links between more closely related nodes (according to the topology used) over less closely related nodes. To measure the modular-ness of a network as a *whole*, a new metric called $Q_2$ is introduced in section 2.

Experiments in section 3 using test problems generated with the method in section 2 support the following argument: (i) modularity increases difficulty for hill climbers, (ii) modularity decreases difficulty for genetic algorithms, (iii) and thus modularity increases the performance advantage of genetic algorithms over hill climbers, (iv) but modularity can also increase difficulty for genetic algorithms if mutation becomes too ineffectual, (v) further, it is harder for mutation to become ineffectual on test problems with heavy-tailed right-skewed connectivity, and (vi) thus degree distribution type does effect evolutionary algorithm performance. We conclude that when modularized, test problems with broadly connected constraint networks are easier for both hill climbers and genetic algorithms to solve. This conclusion is explained in section 4 by the role of richly connected nodes or *hubs* which is the main structural difference between bell-shaped (Poisson for large N) and broad connectivity networks.

## 2. TEST PROBLEM GENERATION

A test problem comprises a set of *iff* (if-and-only-if) constraints defined on a set of variables which are arranged as a string. Each *iff* constraint is defined between a pair of variables; and an *iff* constraint between two variables *i* and *j* is satisfied if and only if *i* and *j* hold the same values, i.e. $i = j$. Fitness of a string *S* is measured by counting the number of *iff* constraints satisfied by *S*, i.e. $F(S) = \sum_i w_i c_i(S)$ where $c_i(S) = 1$ if $S$ satisfies *iff* constraint $c_i$ and 0 otherwise. Each *iff* constraint may be associated with a weight value $w_i$. In this paper, all weight values are 1.0 and all variables take binary values only, i.e. $S = \{0, 1\}^N$, and N=200. Solving a test problem involves finding a string that maximizes the number of satisfied *iff* constraints.

A test problem can be viewed as a network (graph) of nodes and links (edges) where each node represents a problem variable and each link denotes an *iff* constraint. The algorithm described in this section is used to generate such networks. The networks generated are simple graphs, i.e. unweighted, undirected, and have no loops (self-edges) and no multiple edges. This type of network is commonly studied in the field of complex networks.

### Algorithm

1. Create a random graph with N nodes and a given *node degree list (ndl)*. Let the resultant graph be $G_0$.
2. Randomize $G_0$ by exchanging or *switching* pairs of edges selected uniformly at random. Let the resultant graph be $G_r$.
3. Generate *T*, the *decomposition topology*.
4. Measure $aed(G_r)$, the *average edge distance* for $G_r$ relative to *T*.
5. Modularize $G_r$ by selectively exchanging pairs of edges preferentially selected at random. Edge switching is biased towards increasing edge distances relative to *T*. Let the resultant graph be $G_m$.
6. Measure $aed(G_m)$, the average edge distance for $G_m$ relative to *T*. $aed(G_m) > aed(G_r)$ is interpreted as $G_m$ being more modular than $G_r$. Using $aed(G_m)$ and $aed(G_r)$, measure modular-ness of $G_m$ with the $Q_2$ metric as follows: $1.0 - \frac{aed(G_r)}{aed(G_m)}$.

### Step 1: Node degree lists and random graph $G_0$ creation.

The degree of a node $deg(n)$ is the number of edges adjacent to *n*. A node degree list (*ndl*) enumerates the degree for all nodes in an undirected graph in ascending node label order (assumes all nodes are uniquely labeled and node *i* represents problem variable *i*). Two basic conditions for a well-formed node degree list are: (i) it must sum to an even number, and (ii) all its elements must be positive integers.

The *ndl*s of this paper are labeled 1, 2 and 9 to 14 (section 3). The values of an *ndl* are in the order generated by the random number generator and satisfy an additional condition: (iii) all its elements must be much smaller than the number of nodes N, and at least as large as the minimum node degree $deg_{min}$. To induce the formation of connected graphs so that all nodes of a network belong to a single graph component, we set $deg_{min}$ to three. The modules of a hierarchy need not be isolated from each other, but are interrelated subsystems of a larger encompassing whole [17]. This is why it is preferable that the constraint networks be connected graphs. Further, inter-module links introduces *non-separability* into a problem which in turn increases frustration for evolutionary algorithms [19]. A separable problem can be solved by solving individual parts in isolation and aggregating the solution to produce an optimal whole.

A random graph $G_0$ is created from an *ndl* in the usual manner by picking two distinct nodes uniformly at random and placing a link between them if they are not already connected to each other and the node degree list is preserved.

### Step 2: Randomization of $G_0$.

Using a common procedure in random network formation, pairs of links in a $G_0$ are chosen uniformly at random and exchanged if permissible, to reduce any bias inadvertently introduced into $G_0$ in step 1. Edges are exchanged only if the switch does not introduce loops or multiple-edges. Suppose the pair of original links to be switched is $(p, q)$ and $(r, s)$. Two patterns of exchange are used: $(p, s)$ and $(q, r)$; and $(p, r)$ and $(q, s)$. Attempts at edge switching are made $0.125 \times N(N-1)/2$ times, which is 2,487 for networks in this paper. The most number of links a network in this paper has is 694, so each link would have had a chance to switch.

### Step 3: Decomposition topology *T* creation.

The decomposition topology *T* is used in step 5 to guide the formation of modules in a network. Previously, Clauset et al. [4] proposed the creation of *hierarchical random graphs* (random graphs with hierarchical structure) from a pre-specified topology in the form of a *dendrogram D* (a special kind of binary tree). The leaf nodes of *D* represent graph nodes, while the non-leaf or internal nodes of *D* identify groups of related graph nodes, i.e. modules. Each internal node *r* in *D* is associated with a probability value $p_r$. The probability of connecting two graph nodes *i* and *j* is $p_r$ where *r* is the lowest ancestor node in *D* common to *i* and *j*. The values $p_r$ can be adjusted to favour different types of connections.

We use conventional binary trees, and do not require a set of pre-defined probabilities $\{p_r\}$. The function of $\{p_r\}$ is taken up by edge distance (defined in step 4) and the edge switching condition in step 5 which favours short-range connections over long-range ones where possible (although the reverse or other edge switching condition may be specified). Like the internal nodes of *D*, nodes in *T* help identify modules – these modules are theoretical because it remains to be seen whether the actual network, $G_m$, respects them. *T* carves out modules so that smaller modules are nested within larger modules to form a hierarchy, and there is no overlapping of territory between modules at the same level.

All networks in this paper use the same $T$ which is created as follows: nodes of $G_0$ are arranged by node label in a string, and this string is recursively split into two (almost) equal sized halves at node labeled $x$ until the remaining portion is smaller than some *size ts*, which is set to four here. The set of all $x$ node labels derived from this process forms $T$. Figure 1 depicts $T$ for N = 20 and $ts$ = 4. Nodes of $T$ are denoted *internal nodes*. Edges between internal nodes are *internal edges*, and a path comprised exclusively of internal edges is an *internal path*. All internal paths originate at the root of $T$. Other kinds of trees or structures could be generated for $T$.

**Step 4: Measuring edge distance *ed*, relative to *T*.**

Edge distance is a measure of the relatedness between a node pair. The distance of an edge $e = (x, y)$ is the length of the longest internal path shared by $x$ and $y$. As in a hierarchical random graph where more closely related nodes have lowest common ancestors which are situated lower in $D$ than distantly related nodes, nodes incident on edges with larger edge distance values are more related to one another in the sense that they are more likely to belong to the same module according to $T$.

The following example is with reference to Figure 1. Let $e_1 = (1, 4)$, $e_2 = (5, 4)$ and $e_3 = (4, 17)$. The longest internal paths (defined in step 3) for nodes 1, 4, 5 and 17 respectively are: ⟨10i, 5i, 2i⟩, ⟨10i, 5i, 2i⟩, ⟨10i, 5i, 7i⟩ and ⟨10i, 15i, 17i⟩. Since the longest internal paths for nodes 1 and 4 have two internal edges in common, i.e. (10i, 5i) and (5i, 2i); the edge distance of $e_1$, $ed(e_1)$, is 2. Similarly, $ed(e_2) = 1$, and $ed(e_3) = 0$. Thus, of nodes 1, 5 and 17, node 4 is more related to node 1 than to node 5, and least related to node 17. The lowest common ancestor for nodes 1 and 4 is 2i, which is lower in $T$ than 5i, the lowest common ancestor for nodes 4 and 5.

The average edge distance for a graph $G$, $aed(G)$, is the sum of all edge distances divided by the number of edges in $G$:
$$\frac{1}{M}\sum_{i=0}^{M-1} ed(e_i)\, w_i .$$ M is the number of links, and since networks are unweighted, $w_i = 1.0$ for all $i$.

**Step 5: Modularization of $G_r$.**

The link switching method in step 2, with additional conditions, is used to modularize $G_r$ as follows:

(i)  The *complementary edge distance* (*ced*), for all edges is calculated. The complementary edge distance for an edge $e$ is the longest internal path in $T$ $ed_{max}$, less the edge distance of $e$ plus 1, i.e. $ced(e) = ed_{max} - ed(e) + 1$. The "plus 1" ensures that all edges are included at least once in *all_edges*. A randomized (shuffled) list *all_edges*, is made of all edges according to their *ced* values, i.e. if $ced(e_0) = 2$, then $e_0$ will appear twice in this list.

(ii) Distinct pairs of edges are selected uniformly at random from *all_edges* for exchange in the step (iii). In this way, edges with smaller edge distances linking less related nodes relative to $T$ is preferentially selected for modularization.

(iii) Let $e_1 = (p, q)$ and $e_2 = (r, s)$ be the distinct pair of edges selected in step (ii). Then, the two pairs of alternative edges are $e_3 = (p, r)$ and $e_4 = (q, s)$, and $e_5 = (p, s)$ and $e_6 = (q, r)$. Let the edge distance for edge $e_i$ be $ed_i$. If an edge is a loop or introduces a multiple-edge, its edge distance is -1. Let $ped_{ij}$ be the product of edge distances of a pair of edges $i$ and $j$, i.e. $ped_{ij} = ed_i \times ed_j$. The original edge pair $(e_1, e_2)$ is exchanged with the edge pair that has the larger $ped_{ij}$ value larger than $ped_{12}$. Thus, $(e_1, e_2)$ is switched to $(e_3, e_4)$ only if $ped_{34} > ped_{12}$ and $ped_{34} \geq ped_{56}$, and $(e_1, e_2)$ is switched to $(e_5, e_5)$ only if $ped_{56} > ped_{12}$ and $ped_{56} \geq ped_{34}$.

(iv) If a switch is made in step (iii), the modularization algorithm goes to step (i), otherwise it loops to step (ii).

The modularization algorithm iterations through steps (i) to (iv) $P_g \times [M + N (N-1) / 2]$ times. $P_g$ is a parameter for modularizing networks and it is set to 0.8 here. The number of edges M, is included in the number of times the modularization algorithm is iterated to accommodate graphs with the same number of nodes N, but significantly different number of edges. Increasing the number of times the modularization algorithm is applied on a network need not result in a more modular network because a network's degree distribution in part constraints a network's structural possibilities. For example, node degree lists 13 and 14 have the two largest M amongst all the *ndl*s (Table 2), but their $Q$ and $Q_2$ (Figure 3) values are on the smaller side. This is expectable since nodes with unusually high degree have no choice but to make inter-module links.

The modularization algorithm has the effect of reducing the number of edges with smaller edge distances, and increasing the number of edges with larger edge distances (Figure 2). However, step (iii) does not necessarily favour the preservation of edges with larger edge distances over edges with smaller edge distances, and permits for instance the exchange of a pair of edges with edge distances 1 and 8 with a pair of edges with edge distances 4 and 5. This exchange favours the formation of larger modules over the formation of smaller modules.

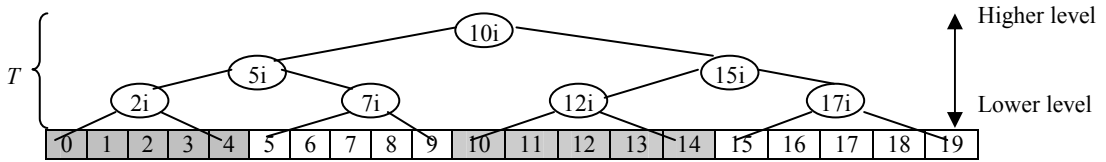

**Figure 1. A decomposition topology** $T$ for N=20 and $ts$ = 4. The internal nodes, i.e. nodes of $T$, are labeled $xi$ to distinguish them from the actual network nodes which are arranged in a row in ascending node label order at the bottom of $T$. Internal nodes identify modules. E.g. internal node 5i identifies the module encompassing nodes 0 to 9, while internal node 2i marks that nodes 0 to 4 belong to a module. The module identified by 2i is organized in $T$ to nest directly within the module identified by 5i.

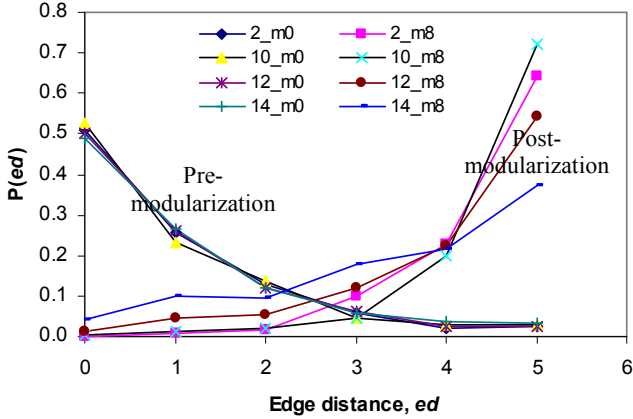

**Figure 2. Change in edge distance distribution due to modularization.**

**Step 6: Measuring modular-ness with $Q_2$.**

It is proposed here that modular-ness of a network $G_B$ be assessed relative to the modular-ness of a comparable network $G_A$ by $Q_2$ as follows: $1.0 - \frac{aed(G_A)}{aed(G_B)}$. $Q_2$ is 0.0 if $G_B$ is as (not) modular as $G_A$, i.e. $aed(G_B) = aed(G_A)$, $Q_2$ is > 0.0 if $G_B$ is more modular than $G_A$, i.e. $aed(G_B) > aed(G_A)$, and $Q_2$ is < 0.0 if $G_B$ is less modular than $G_A$, i.e. $aed(G_B) < aed(G_A)$. In this paper, $G_r$ is $G_A$, and $G_m$ is $G_B$. In its usage here, $Q_2$ compares what is presumed to be a modularized graph with its previous random and therefore likely less modular version. It is possible, as in a fully connected simple graph, that there are no alternative simple graph configurations. In this case, $Q_2$ will be 0.0, which is appropriate since all nodes in a fully connected graph belong to the one same module and thus does not exhibit modularity.

To verify that the modularization algorithm in step 5 does indeed modularize a network, $Q$ values for the first three highest levels are taken before and after step 5. The $Q$ metric was introduced in [14] as a measure of modularity given a certain division of a network. $Q = \frac{s^T B s}{2m}$; where $s$ is a column vector of ±1 elements representing a particular division of a network into two candidate modules, $s^T$ is the transpose of $s$, and $B$ is a real symmetric matrix called the *modularity matrix* with elements $B_{ij} = A_{ij} - \frac{k_i k_j}{2m}$. $A$ is the adjacency matrix for the network where $A_{ij} = e$ means $e$ links exists between nodes $i$ and $j$, $k_i$ and $k_j$ are the respective degrees of nodes $i$ and $j$, and $m$ is the total number of links in the network. Elements of $B$ reflect the statistical surprising-ness of links relative to what could be expected by random chance. A positive (negative) $Q$ value indicates that a network has fewer (more) links than expected between its two divisions as delineated by $s$. Each row and each column of $B$ sum to 0 which assures the existence of an all ones eigenvector with an eigenvalue of zero. A network is indivisible when no other $s$ but the all ones vector produces a non-negative $Q$ value. Indivisible networks have a $Q$ value of 0.0. Optimally divided networks have a $Q$ value of 1.0.

Since $T$s in this paper are binary trees which subdivide modules into two more or less equal halves, the $s$ vectors for $Q$ follow suit. For example, to calculate the highest level $Q$ value for a network with $T$ in Figure 1, $s$ has 20 elements with +1 in its top half and -1 in its bottom half. There are two $Q$ values at the second highest level. The $Q$ value for one module is calculated for nodes 0 to 9, and its $s$ has 10 elements with +1 in its top half and -1 in its bottom half. The $Q$ value for the other module is similarly calculated for nodes 10 to 19. Table 1 gives a sample of the $Q$ and $Q_2$ values before modularization (m0) and after modularization (m8) for networks generated from four different *ndl*s. The degree distribution curve for *ndl* 2 is bell-shaped while it is right-skewed and heavy-tailed for the other three *ndl*s (section 3). Prior to modularization, the $Q$ values are negative and close to 0.0000 and the $Q_2$ values by definition is 0.0000. After the modularization algorithm in step 5 is applied with $P_g = 0.8$, the $Q$ values for the first three highest levels increase significantly towards 1.0. The average edge distance (*aed*) values for modularized networks are at least 3.5 times that of non-modularized networks. As such, $Q_2$ values of the modularized networks rise significantly above 0.0000 (Figure 3). In short, the networks do become more modular after step 5, and the $Q_2$ measure does indicate increase in modular-ness of a network.

However, while the correlation between the $Q$ values for the highest level and corresponding $Q_2$ values is strong (0.8487 for the test set in section 3), the two measures need not necessarily rank networks by modular-ness in the same order. This reflects the semantic difference between $Q$ and $Q_2$. While $Q$ measures modular-ness of a network with respect to a particular division of the network at a single level, $Q_2$ considers the modular-ness of a network in its entirety with respect to a particular decomposition topology $T$. N1 = {(0, 1), (0, 2), (0, 3), (0, 4), (4, 6), (4, 5), (4, 7)} and N2 = {(0, 1), (0, 2), (0, 4), (2, 3), (4, 5), (4, 6), (6, 7)} where $(u, v)$ represents a link between nodes $u$ and $v$ are two networks of the same size (N=8, M=7). Assuming $T$ is a perfect binary tree with $ts = 2$, N1 and N2 have the same $Q$ value (0.7143) for the highest level, but compared with the same random graph N1's $Q_2$ value is smaller than N2's. Hence, the $Q$ and $Q_2$ metrics are distinguishable from each other.

**Table 1. Network modular-ness pre- (m0) and post- (m8) modularization.**

| N = 200 | $Q$ values for the first three highest levels | | | | | | | *aed* | $Q_2$ |
|---|---|---|---|---|---|---|---|---|---|
| 2_m0  | -0.0232 | -0.0069 | -0.1067  | -0.1940 | -0.0900 | -0.2407 | 0.0220  | 0.8940 | 0.0000 |
| 2_m8  | 0.9966  | 0.9666  | 0.9934   | 0.9725  | 0.9690  | 0.9339  | 0.9731  | 4.4669 | 0.7999 |
| 10_m0 | -0.0554 | 0.0893  | -0.0750  | -0.5936 | 0.1023  | -0.1478 | -0.4025 | 0.9014 | 0.0000 |
| 10_m8 | 0.9905  | 0.9715  | 0.9662   | 0.9268  | 0.9611  | 0.9456  | 0.9799  | 4.5757 | 0.8030 |
| 12_m0 | -0.0097 | -0.11097| -0.08526 | -0.5147 | -0.1736 | 0.0636  | -0.0494 | 0.9219 | 0.0000 |
| 12_m8 | 0.9732  | 0.8371  | 0.9555   | 0.6585  | 0.9636  | 0.9636  | 0.9518  | 4.1295 | 0.7768 |
| 14_m0 | 0.0200  | 0.0394  | -0.1286  | 0.2849  | -1.0149 | -0.8272 | -0.1720 | 0.9896 | 0.0000 |
| 14_m8 | 0.9162  | 0.8118  | 0.7638   | 0.8887  | 0.6762  | 0.7780  | 0.5917  | 3.5460 | 0.7209 |

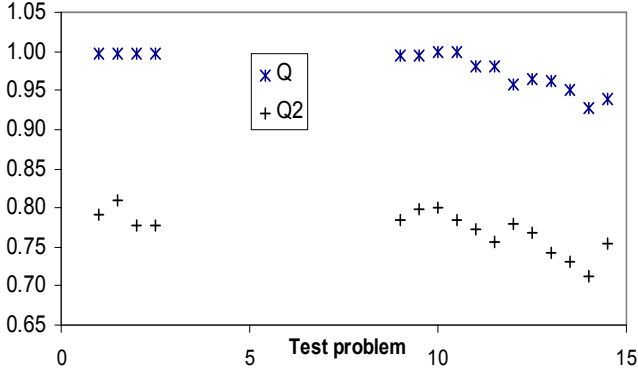

**Figure 3.** $Q$ and $Q_2$ values for modularized test problems.

A *clustering coefficient spectrum* that is inversely related with node degree is interpreted as indicative of hierarchical organization [16]. A network's clustering coefficient spectrum $C(k)$ values are obtained by averaging the clustering coefficient of node $i$ $C_i$ for all $i$ with degree $k$. $C_i$ is the ratio of actual to possible links amongst a set of nodes: $C_i = \dfrac{2E_i}{k_i(k_i-1)}$. $E_i$ is the number of links between node $i$'s $k$ neighbors, and $k(k-1)/2$ is the number of possible (undirected single) links between $k$ nodes [20]. After modularization, $C(k)$ values for broad connectivity networks (*ndl*s 9 to 14) become significantly more inversely related with $k$, while $C(k)$ values for random connectivity networks (*ndl*s 1 and 2) show almost uniform increases independent of $k$. This illustrates that random networks can become hierarchically organized after applying the modularization step following the binary tree decomposition topology $T$ but that the level of hierarchical organization achieved depends also on degree distribution type.

## 3. EXPERIMENTS

The objective of the experiments is to explore the effect of modularity and degree distribution on evolutionary algorithm performance. The experiments are carried out on test problems with constraint networks generated with the method in section 2. The number of nodes N in a constraint network is also the size of a test problem. N is 200 for all test problems. The total number of links M is also the fitness value of an optimal genotype.

The node degree lists (*ndl*s) are produced by rounding the values generated by the *randht.m* procedure (version 1.0.2) provided online by A. Clauset. Two different kinds of distributions are used: normal (*ndl*s 1 and 2) and power-law with degree exponents 4.0 (*ndl*s 9 and 10), 3.0 (*ndl*s 11 and 12) and 2.6 (*ndl*s 13 and 14). The characteristics of the *ndl*s are summarized in Table 2. The degree distribution of *ndl*s 1 and 2 have little to no skew, mean = median = mode. The degree distribution curves of the other *ndl*s are right-skewed, mean ≥ median ≥ mode. The standard deviations noticeably increase going down the list of *ndl*s in Table 2. Figure 4 depicts the complementary cumulative distribution function (CDF): $P(x) = \Pr(X \geq x)$ of *ndl*s 9 to 14 on a double log-scale. Whether these distributions are best characterized by power-laws is less important than that they are heavy-tailed and right-skewed. The doubly logarithmic scale is used as it is a convenient form to depict the distributions. Power-law identification for the *ndl*s is hampered by the fact that N=200 which is small compared to the hundreds of nodes in real-world scale-free networks. $deg_{min}$ which is 3, is also smaller than 5 which increases the error associated with out method of generating the *ndl*s [5]. Nevertheless, using a larger N would substantially increase simulation time without necessarily producing more relevant insight. The *ndl*s may be produced by network growing models such as the many variations of the preferential attachment model [2] and other generalized random graph models [12]. This is a logical next step.

**Table 2. Node degree list summary statistics**

| *ndl* | Min | Max | Mean | Std. dev | Mod | Median | M |
|---|---|---|---|---|---|---|---|
| 1 | 3 | 9 | 6.05 | 1.0786 | 6 | 6 | 605 |
| 2 | 3 | 9 | 6.04 | 1.0459 | 6 | 6 | 604 |
| 9 | 3 | 19 | 4.60 | 2.2573 | 3 | 4 | 460 |
| 10 | 3 | 15 | 4.36 | 1.8595 | 3 | 4 | 436 |
| 11 | 3 | 36 | 5.74 | 4.3006 | 4 | 4 | 574 |
| 12 | 3 | 53 | 5.25 | 4.3822 | 3 | 4 | 525 |
| 13 | 3 | 50 | 6.94 | 7.0673 | 3 | 5 | 694 |
| 14 | 3 | 68 | 6.74 | 7.6265 | 4 | 5 | 674 |

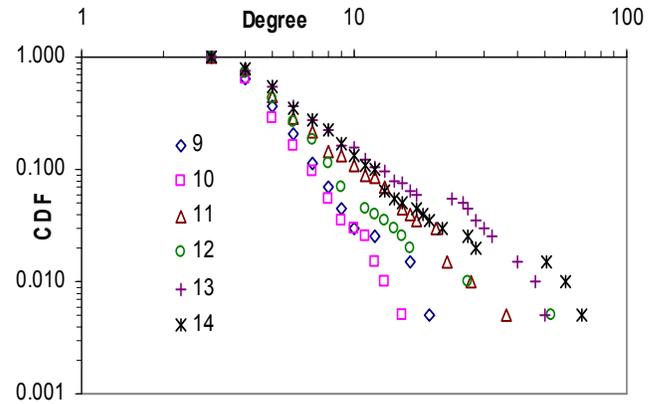

**Figure 4.** The complementary cumulative distributions (CDF) of *ndl*s 9 to 14 on a log-log plot.

Solutions to test problems are represented in the evolutionary algorithms as linear genotypes. Two hill climbing algorithms (HC) are used: (i) the Random Mutation Hill Climber (RMHC) [7] which is implemented as mutating $z$ genes chosen uniformly at random with replacement from a genotype at one time, and (ii) the Macro-Mutation Hill Climber (MMHC) [10] which is implemented as mutating $z$ consecutively located genes in a genotype from left to right at one time, starting from a randomly chosen loci and the genotype is treated as a ring. The integer $z$ is chosen uniformly at random from $[1, P_m \times N]$ where $P_m$ is a parameter controlling the mutation rate. $P_m$ values in the experiments are 0.0625 and 0.125. Both RMHC and MMHC use bit-flip mutation and replace the current genotype (parent) with its mutant offspring if the mutant is as fit as or fitter than its parent. The genetic algorithm (GA) used is a steady-state GA that does uniform random parent pair selection, applies with $P_x$ probability two-point crossover (which allows for one-point crossover) to a parent pair, and applies with 1.0 - $P_x$ probability random bit-flip mutation (same as RMHC) to each of the parent genotypes. A crossover offspring replaces a parent only if it is fitter than both its parents. A mutation offspring replaces its parent if it is not less fit than its parent. Details of this GA have been published elsewhere by the author. Population size (PS) is 100, crossover probability ($P_x$) is 0.25 and mutation rate ($P_m$) is 0.0625 or 0.125.

The evolutionary algorithms are run until an optimal solution is found or 250,000 function evaluations have been made. Two random networks are created for each *ndl* using a different random number seed each time, and the modularization process is applied to each of these networks. Each one of these networks makes up a test problem instance, e.g. test problem 1 comprises two test problem instances each with its own constraint network created using *ndl* 1. The evolutionary algorithms are run 10 times each with a different random number seed each time, on each test problem instance.

## 4. RESULTS AND DISCUSSION

Table 3 summarizes for different $P_m$, the success rates of RMHC, MMHC and GA on random networks (m0) and their modularized selves (m8). Error bars in figures mark 95% confidence intervals. Some points are positioned on the x-axis with a small positive offset to avoid overlapping. The results support the following assertions:

(i) Modularity increases difficulty for hill climbers, more so for random mutation than macro-mutation. The success rates of both RMHC and MMHC drop significantly when the networks are modularized, but the decline is steeper for RMHC than MMHC. Increasing $P_m$ increased the success rate of MMHC on modularized networks, but at the cost of significant increases in average number of function evaluations (Figure 5). Modularity creates fitness saddles in mutation-only fitness landscapes [19]. Increasing the mutation rate for macro-mutation reduces the widths of these fitness saddles, thus making them easier to overcome. However, increasing $P_m$ did little to improve the performance of RMHC. Genotypes at end of unsuccessful RMHC runs show large all-zeroes segments interspersed with large all-ones segments, revealing unbridgeable fitness saddles.

(ii) Modularity decreases difficulty for genetic algorithms. The success rate of GA improves by about 10% (Table 3), and the average number of evaluations to solve a problem (MFPT) decreases significantly when the networks are modularized (Figure 6). The GA also performs better– its success rate is higher (Table 3), and its MFPT is not significantly larger (Figure 6) – when a smaller mutation rate is used. On non-modularized test problems 9 and 10, the GA can be 100% successful, but the evaluation time needs to be almost doubled. The 95% confidence interval of MFPT for these runs range between 255,850 and 387,750. Modularization favours links between more related nodes and because tight genetic linkage is respected, more related nodes tend to be located nearer to each other on the linear genotype. This is helpful towards crossover success in the GA. The modular organization imposes a restriction on the probable defining lengths of fit schemata. The defining length of fit schemata is inversely related to their conservation, and the incremental construction of larger order fit schemata is central to good GA performance according to the building-block hypothesis [8]. Figure 7 compares the distribution of defining lengths for fit schemata of order 2 (edge spans), i.e. the number of possible crossover cut-off points between two inter-dependent genes. Fit order-2 schemata are more likely to have shorter defining lengths in modularized (m8) than in non-modularized (m0) test problems.

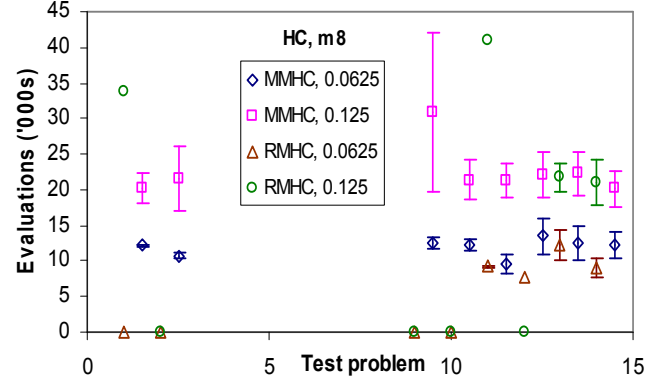

**Figure 5. Average evaluations of successful* HC runs on modularized (m8) test problems.** Mutation rates ($P_m$) are 0.0625 and 0.125. *Evaluations = 0 means 0% success rate.

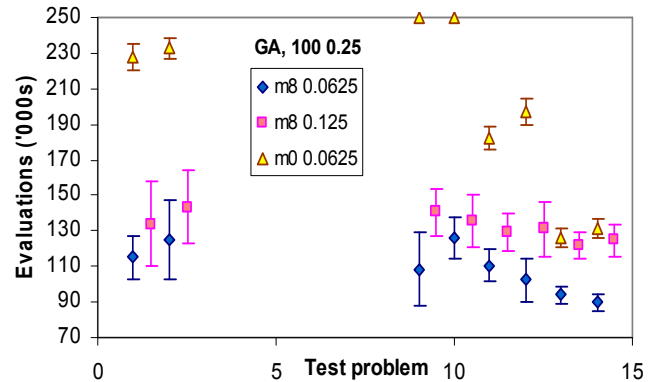

**Figure 6. Average evaluations of successful* GA runs on non-modularized (m0) and modularized (m8) test problems.** Population size is 100, crossover probability ($P_x$) is 0.25 and mutation rates ($P_m$) are 0.0625 and 0.125. *Evaluations = 250,000 means 0% success rate.

(iii) Modularity increases the performance advantage of genetic algorithms over hill climbers. On non-modularized test problems, the hill climbers are more successful and more efficient than the GA. In contrast to Figure 6 which starts at 70,000 on the y-axis, all hill climbing runs completed successfully in less than 30,000 function evaluations. On modularized test problems, the GA is more successful than the hill climbers, particularly RMHC.

**Table 3. Number of successful runs out of a total of 20 runs for each test problem**

| Test problem (N = 200) | m0 | | | m8 | | | | | |
|---|---|---|---|---|---|---|---|---|---|
| | RMHC | MMHC | GA | RMHC | | MMHC | | GA | |
| | 0.0625 | 0.0625 | 0.0625 | 0.0625 | 0.125 | 0.0625 | 0.125 | 0.0625 | 0.125 |
| 1 + 2 | 40 | 40 | 25 | 0 | 1 | 5 | 21 | 23 | 13 |
| 9 + 10 | 40 | 38 | 0 | 0 | 0 | 7 | 25 | 23 | 13 |
| 11 + 12 | 40 | 39 | 40 | 3 | 1 | 20 | 30 | 36 | 36 |
| 13 + 14 | 40 | 40 | 40 | 13 | 14 | 28 | 30 | 40 | 40 |
| Success rate | 100% | 98.125% | 65.625% | 10% | 10% | 37.5% | 66.25% | 76.25% | 63.75% |

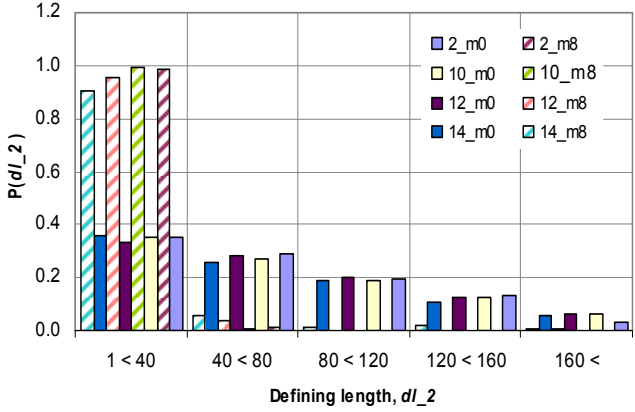

**Figure 7. Changes in the distribution of defining lengths of fit order-2 schemata due to modularization.**

(iv) Modularity can increase difficulty for genetic algorithms. Genetic algorithms, as do natural evolution, rely on random mutation to supply genetic diversity. Since modularity makes mutation less effective (point i) and mutant offspring less able to compete for survival in a steady-state GA population, increases in modularity can be too much of a good thing for genetic algorithms. Ineffectual mutation in a finite population increases the likelihood of gene fixation and if this convergence is unfavourable to the evolution of an optimal solution, the GA prematurely converges. Figure 8 gives the end-of-run *genome convergence ratio* (GCR) averaged over all runs for modularized test problems. Most of these values are significantly higher than corresponding GCR values for non-modularized test problems which were all 0.0000. GCR is the fraction of genes that have converged (all have the same value) in a population at a point in time. While genetic convergence is useful for focusing the search efforts of a genetic algorithm, convergence which is too rapid or of the wrong sort hampers a genetic algorithm from exploring more fertile areas of a fitness landscape. In short, modularity is a double-edged sword for genetic algorithms.

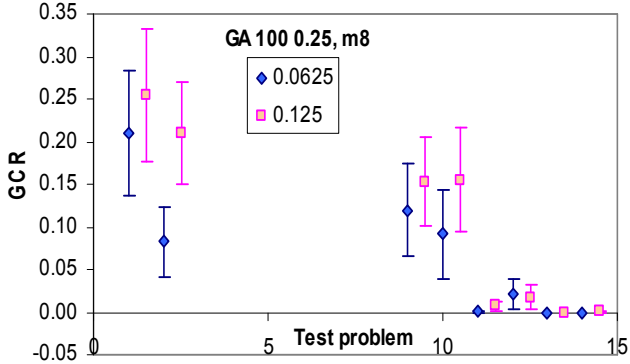

**Figure 8. Genome convergence ratio (GCR) at the end of all GA runs on modularized problems (m8) for $P_m$ = 0.0625 and 0.125 (PS = 100 and Px = 0.25).**

(v) It is harder for mutation to become ineffectual on modular test problems with broad connectivity. RMHC and MMHC are more successful on modularized test problems 11 to 14 than on the other four test problems (Table 3). GCR values at end of modularized test problems 11 to 14 are also significantly smaller than for the other four test problems (Figure 8). The degree distributions of test problems (*ndl*s) 11 to 14 have fatter tails than the other four test problems. Compared with *ndl*s 1 and 2, *ndl*s 11 to 14 have a small number of *hub*s or nodes with significantly more links, and compared with *ndl*s 9 and 10, the hubs of *ndl*s 11 to 14 are more richly connected (Table 2). Hubs play an influential role on processes taking place on networks, e.g. speed of information spread and resilience of network connectedness [Newman, 2003]. Likewise, hubs can influence the effectiveness of mutation on modular test problems.

Modularity poses a problem for mutation-only algorithms because it induces the emergence of domain walls [11]. Further, with no strong external beacon to guide adaptation, individual modules are free to adapt to any global optima. But because the test problems are non-separable, mere aggregation of optimal modules need not produce an optimal solution. Lack of coordination between the adaptation efforts of different modules can also cause problems for a GA. Although crossover helps a GA eliminate domain walls, it can only do so if modules with the right genetic material are available for exchange. The inability of a GA to create the right genetic modules because mutation is too weak is the *synchronization problem* [18].

Hubs can help both hill climbers and the GA solve the test problems by helping to coordinate the adaptation efforts of different modules within a genotype, and by quickly disseminating information about their values to other non-hub genes within a genotype so that the non-hub genes can adapt before it is not possible for them to do so (the fitness saddle becomes too wide for the mutation operator). But hubs can do this only if they are synchronized themselves in the sense of having the right values to create an optimal solution, and they are central to inter-node communication. These two conditions are satisfied to a greater extent in networks with broad connectivity.

Compared with the other test problems, the modularized constraint networks of test problems 11 to 14 have smaller diameters and shorter average shortest path lengths (SPL) (Figure 9 top). This speeds up communication between genes in a genotype. More importantly, the SPLs amongst hubs are much shorter in modularized broadly connected networks (Figure 9 bottom) and this facilitates synchronization amongst hubs. Hubs in modularized broadly connected networks also occupy a much more central position in inter-node communication on a network. After modularization, the positive correlation between node degree and *centrality* declines more substantially for test problems 1, 2, 9 and 10 than for the other test problems (Figure 10). Centrality of a node measures the number of shortest paths between other nodes that traverses the node [13]. It is also important that once hubs have the right values to evolve an optimal solution, they are stable so that a consistent message is received by the other non-hubs nodes. This condition is again better satisfied in networks with broad connectivity. Figure 11 shows that hubs mutate less frequently than non-hub nodes in test problem 13 which is understandable since changing the value of a hub gene can cause large changes to genotype fitness. Modularization does not alter the general pattern of successful mutation frequency of genes.

To summarize, hubs exert a coordinating, directing and stabilizing force over the adaptation of a genotype. This can be helpful for conducting search in frustrating fitness landscapes as demonstrated in this paper. We are our study to other network structural characteristics to understand their inter-related influences on evolution and evolvability of complex systems.

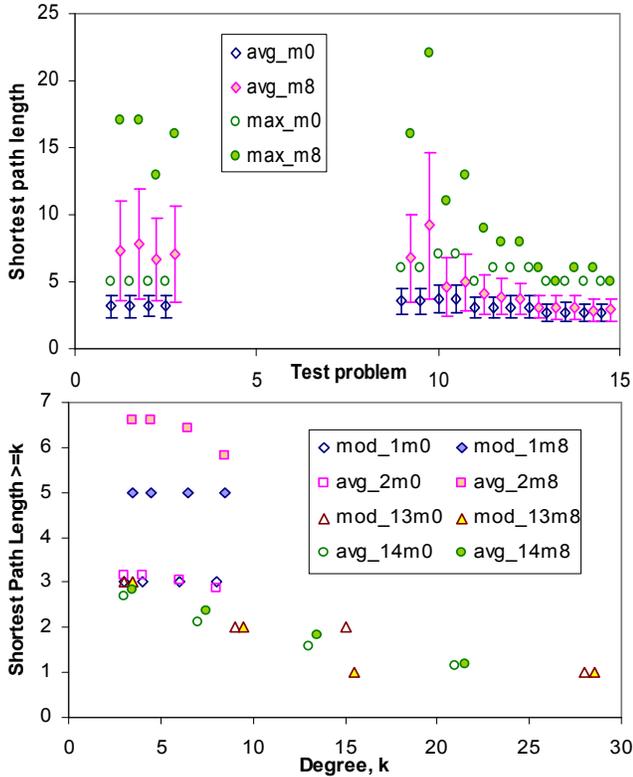

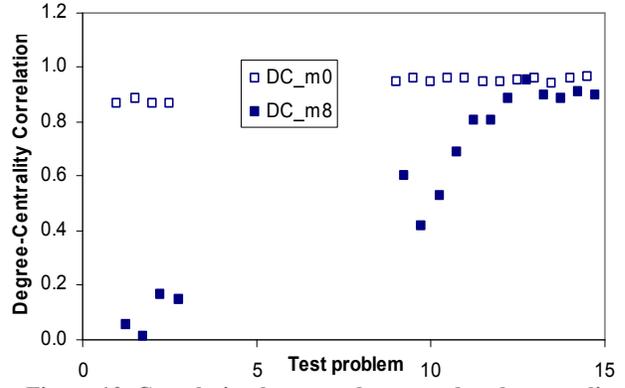

Figure 10. Correlation between degree and node centrality.

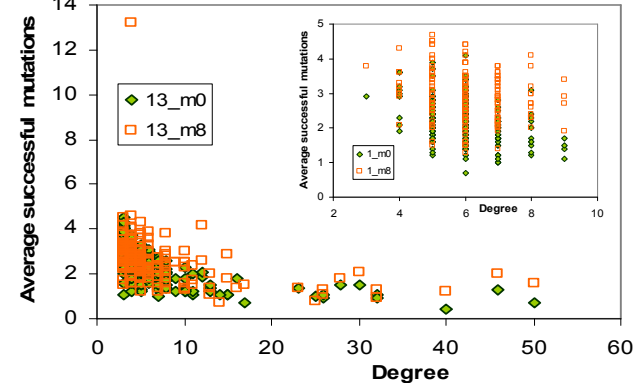

Figure 9. Top: Average and maximum shortest path lengths (SPL) of non-modularized (m0) and modularized (m8) test problems. Bottom: Average and most frequently occurring SPL amongst nodes with degree $k$ or higher.

Figure 11. Successful bit-flip mutations averaged over 10 RMHC $P_m = 0.0625$ runs for non-modularized (m0) and modularized (m8) test problem 1 (insert) and 13.